\pdfoutput=1

\documentclass[11pt]{article}

\usepackage[final]{acl}

\usepackage{times}
\usepackage{latexsym}

\usepackage[T1]{fontenc}

\usepackage[utf8]{inputenc}

\usepackage{microtype}

\usepackage{inconsolata}

\usepackage{graphicx}
\usepackage{booktabs}
\usepackage{multicol}
\usepackage{multirow}

\usepackage[accsupp]{axessibility}  %

\usepackage{dsfont}
\usepackage{xcolor}
\usepackage{graphicx}
\graphicspath{figures}

\usepackage{algorithm}
\usepackage{algorithmicx}
\usepackage{algpseudocode}
\usepackage{booktabs}
\usepackage{adjustbox}

\usepackage{subcaption}

\usepackage{mathtools}

\newcommand{\RR}{\mathbb{R}}

\newcommand{\argmin}{\mathop{\mathrm{arg\,min}}}

\newcommand{\Var}{{\mathop{\mathrm{Var}}}}

\newcommand{\mse}{\mathrm{MSE}}

\renewcommand{\bar}{\overline}

\renewcommand{\tilde}{\widetilde}

\renewcommand{\hat}{\widehat}

\usepackage{./macros/macros_03_symbols}
\usepackage{./macros/macros_04_propernames}

\newcommand{\bias}{\mathop{\mathsf{bias}}}
\newcommand{\var}{\mathop{\mathsf{var}}}
\renewcommand{\mse}{\mathop{\mathsf{mse}}}

\newcommand{\reg}{SR\xspace}

\newcommand{\overbar}[1]{\mkern 1.5mu\overline{\mkern-1.5mu#1\mkern-1.5mu}\mkern 1.5mu}

\renewcommand{\bar}{\overbar}

\usepackage[font=small,skip=5pt]{caption}

\usepackage[xspace]{ellipsis}
\usepackage{comment}

\usepackage{crossreftools}
\usepackage[
    textsize = scriptsize,
]{todonotes}

\usepackage{cleveref}

\usepackage{tabularx}

\usepackage{enumitem}

\newcommand{\kmeans}{$k$-means\xspace}

\renewcommand{\epsilon}{\varepsilon}

\title{Precise Model Benchmarking with Only a Few Observations}

\author{Riccardo Fogliato\thanks{Corresponding author} \\ %
  Amazon Web Services \\
  \texttt{fogliato@amazon.com}
  \And
  Pratik Patil \\
  University of California, Berkeley \\
  \texttt{pratikpatil@berkeley.edu} 
  \AND
  Nil-Jana Akpinar \\
  Amazon Web Services \\
  \texttt{nakpinar@amazon.com}
  \And
  Mathew Monfort \\
  Amazon Web Services \\
  \texttt{monfortm@amazon.com}
  }

\begin{document}

\maketitle

\begin{abstract}
How can we precisely estimate a large language model's (LLM) accuracy on questions belonging to a specific topic within a larger question-answering dataset? 
The standard direct estimator, which averages the model's accuracy on the questions in each subgroup, may exhibit high variance for subgroups (topics) with small sample sizes.
Synthetic regression modeling, which leverages the model's accuracy on questions about other topics, may yield biased estimates that are too unreliable for large subgroups.
We prescribe a simple yet effective solution: an empirical Bayes (EB) estimator that balances direct and regression estimates for each subgroup separately, improving the precision of subgroup-level estimates of model performance.
Our experiments on multiple datasets show that this approach consistently provides more precise estimates of the LLM performance compared to the direct and regression approaches, achieving substantial reductions 
in the mean squared error.
Confidence intervals for EB estimates also have near-nominal coverage and are narrower compared to those for the direct estimator. 
Additional experiments on tabular and vision data validate the benefits of this EB approach.
\end{abstract}

\section{Introduction}
\label{sec:introduction}

Accurate evaluation of large language models (LLMs) is crucial for identifying their strengths and weaknesses.
While broad topics like math or history often have lots of data available for testing, specific topics may lack suitable test data.
For example, we may not be able to collect queries for niche subjects, such as a particular historical event, legal terms from specific jurisdictions, rare medical conditions, or regional dialects. Consequently, we may fail to reliably assess the LLM’s understanding of these topics.
When gathering more data is not an option, practitioners must rely on existing datasets to measure the model’s performance.

\begin{figure}[!t]
    \includegraphics[width=0.99\columnwidth]{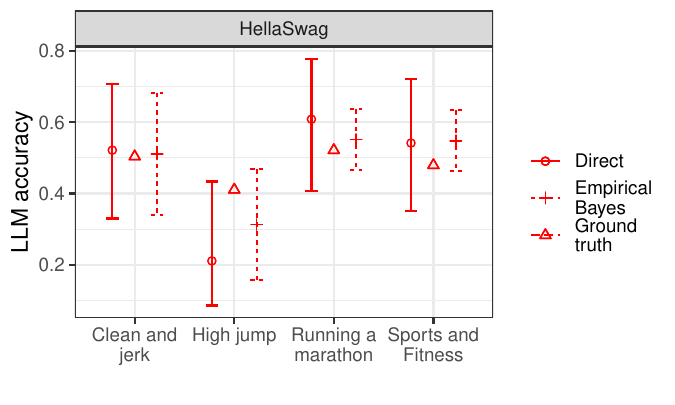}
    \caption{
        \textbf{Estimates of LLM accuracy and their 95\% confidence intervals} for predictions made by Gemma-2b across various subgroups on a subset of HellaSwag. 
        The empirical Bayes estimates have precision similar to the direct estimator for some subgroups (e.g., clean and jerk) and higher precision for others (e.g., high jump). 
        This approach also provides tighter confidence intervals for the estimates; e.g., see running and sports topics.
        }
        \label{fig:results_initial}
        \vspace{-1em}
\end{figure}

To illustrate this problem, consider a randomly sampled subset of HellaSwag \cite{zellers2019hellaswag}, a dataset with multiple-choice questions on various topics or subgroups; e.g., see \Cref{fig:results_initial}.
Our goal is to precisely estimate the accuracy of the answers given by the LLM in each subgroup using only the limited data available.
Traditionally, we estimate model performance via a direct estimator (DT), which computes the average accuracy in each subgroup separately.
However, when a subgroup has only a few questions, the DT estimates can exhibit high variance and become unreliable.

One may expect the performance of the LLM on related subgroups to be associated. 
Synthetic estimation aims to exploit this relationship via regression modeling (\reg) \citep{rao2015small}. 
For example, we can incorporate the LLM's knowledge of sports to aid in estimating its accuracy on questions about running. 
The resulting estimates will generally exhibit less variance compared to direct estimation but may be biased, and thus they may be imprecise for subgroups with large sizes. 

Empirical Bayes (EB) approaches combine DT and \reg estimates, adjusting the contribution of each for every subgroup separately \citep{efron1977stein}. 
In theory, they hold the promise to improve the precision of these baselines \citep{ignatiadis2019covariate} and allow for the construction of confidence intervals \citep{armstrong2022robust}; e.g., see again \Cref{fig:results_initial}. 
However, to the best of our knowledge, an extensive empirical evaluation of the validity of this approach for precise model benchmarking is missing. 

In this work, we empirically assess the precision of EB estimates of LLM performance on subgroups (domains, tasks, topics, etc.) across multiple benchmark datasets.
Our results in \Cref{sec:results} show that EB estimates are consistently more precise than those of DT and \reg. 
Their confidence intervals have good coverage and substantially smaller widths than those of DT. 
Additional experiments on vision and tabular data validate these findings (\Cref{sec:add_additionalexp}).  

\section{Problem Setup}
\label{sec:problem_setup}

Consider a dataset $\mathcal{D} = \{(X_g,Z_g)\}_{g=1}^G$, where $X_g \in \mathcal{X}$ represents the features of subgroup $g$ and $Z_g$ indicates the performance measure for that subgroup.
For example, $X_g$ could be the description of the topic of the questions in the subgroup, while $Z_g$ could denote the LLM's average empirical classification accuracy or Brier score on such questions.
Our goal is to use $\mathcal{D}$ to obtain estimates of subgroup performances $\{\mu_g\}_{g=1}^G$ as accurate as if we had infinite data for each subgroup. 
Formally, we define $\mu_g := \lim_{n_g\to \infty} Z_g$, where $n_g$ denotes the subgroup size.
The objective is to find estimators $\{\hat{\mu}_g\}_{g=1}^G$ that minimize the average mean squared error (MSE) across all subgroups:
\begin{equation}
    \label{eq:evaluation_metric}
     \frac{1}{G}\sum_{g=1}^G \mse(\hat{\mu}_g) = \frac{1}{G}\sum_{g=1}^G \mathbb{E}[(\hat{\mu}_g - \mu_g)^2].
\end{equation}
We say that estimators with lower average MSE are more \emph{precise}.
By standard arguments, the MSE for each subgroup $g$ decomposes into bias and variance components: $\mse(\hat{\mu}_g) = {\bias}^2(\hat{\mu}_g) + \var(\hat{\mu}_g)$, where $\bias(\hat{\mu}_g) := \mathbb{E}[\hat{\mu}_g] - \mu_g$ and $\var(\hat{\mu}_g) := \mathbb{E}[(\hat{\mu}_g - \mathbb{E}[\hat{\mu}_g])^2]$.
In the following, we consider three estimation methods.

\section{Methods}
\label{sec:methods}

\paragraph{Direct estimator (DT).\hspace{-0.5em}} 
The standard approach for estimating $\mu_g$ is the direct estimator, where we use $\hat{\mu}_g = Z_g$ which is the subgroup-conditional empirical average. 
This estimator is unbiased and its MSE is equal to its variance, i.e., $\mse(\hat{\mu}_g) = \var(\hat{\mu}_g)$. 
When the subgroup size $n_g$ is large, this variance is typically small, resulting in a precise estimate. 
However, for small $n_g$, DT suffers from high variance, leading to less reliable estimate. 
To quantify the uncertainty of these estimates, we use Wilson score intervals for binomial proportions (e.g., binary accuracy) \citep{brown2001interval} and Student's t-intervals for other continuous outcomes (e.g., F1 score).

\paragraph{Synthetic regression (\reg).\hspace{-0.5em}}
Synthetic regression estimators leverage information from related subgroups by learning a regression function $f(X_g) = \mathbb{E}[Z_g \,|\, X_g]$. 
This is a common approach in the small area estimation literature \citep{rao2015small}.
The estimator is then given by $\hat{\mu}_g = \hat{f}(X_g)$, where $\hat{f}$ is the fitted regression model. 
We use XGBoost with cross validation \citep{chen2016xgboost}, 
a flexible method that reduces variance through regularization. As features, we employ the text embeddings of $X_g$, LLM confidence scores, as well as task- and model-specific intercepts; other approaches are possible, see \Cref{sec:additional-details-methods} for more details. 
While $\hat{f}(X_g)$ may introduce bias, the hope is that it significantly reduces variance compared to DT, especially for groups with small $n_g$'s.
This would result in $\var(\hat{\mu_g}) \ll \bias(\hat{\mu_g})$.
\reg provides the greatest improvements in precision over the DT estimator when the regression fit is good and the subgroup metric $Z_g$ has high variance. 

\begin{figure*}[ht]
  \centering
  \includegraphics[width=0.95\textwidth]{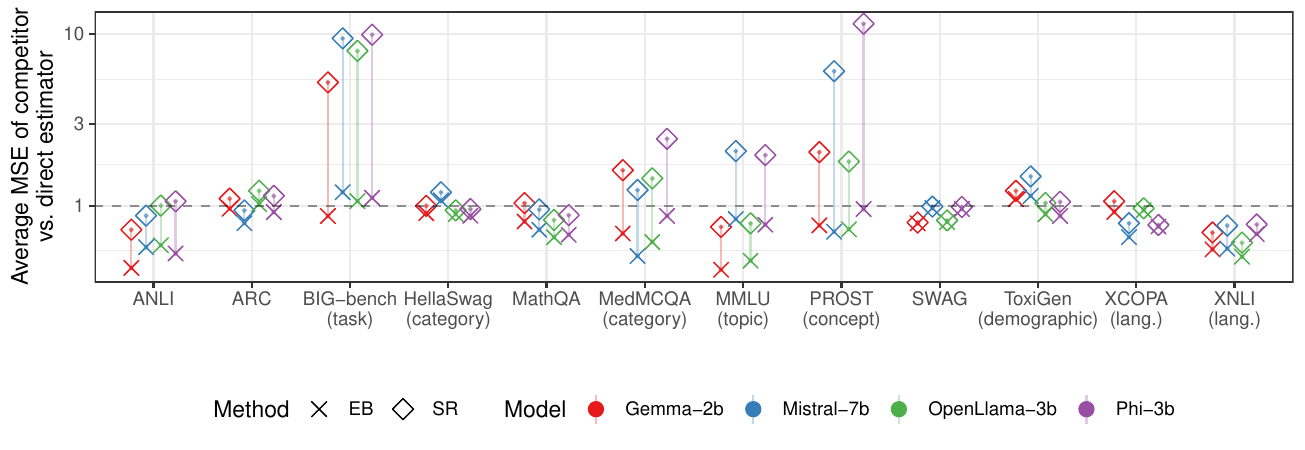}
   \caption{
       \textbf{Comparison of methods to estimate the accuracy of LLMs across datasets.} 
       The plot shows the ratios of the estimates MSEs, obtained using regression (\reg) and empirical Bayes (EB) methods, relative to the direct estimator (DT) for the LLM's accuracies on LLM-task subgroups (in parenthesis when pre-defined). 
       Lower ratio values indicate more accurate estimates compared to DT. 
       EB consistently provides more precise estimates than both \reg and DT across most evaluations.  
   }
   \label{fig:results_main}
   \vspace{-1em}
\end{figure*}

\paragraph{Empirical Bayes estimator (EB).\hspace{-0.5em}}
The empirical Bayes estimator combines the strengths of DT and \reg.
The estimator is formally given by:
\begin{align}\label{eq:eb}
    \hat{\mu}_g=\frac{\hat{\sigma}_g^2}{\hat{\sigma}_g^2+\hat{A}} \cdot \hat{f}(X_g)+ \frac{\hat{A}}{\hat{\sigma}_g^2+\hat{A}} \cdot Z_g,
\end{align}
where $\smash{\hat{\sigma}_g^2=\hat{\Var}(Z_g)}$ is an estimate of the variance of $Z_g$, and $\hat{A}=\{G^{-1}\sum_{g=1}^G [(Z_g-\hat{f}(X_g))^2-\hat{\sigma}_g^2]\}_{+}$. 
The EB estimator dynamically balances the DT and \reg estimates, %
aligning more closely with \reg when $\hat{f}(X_g)$ approximates $\mu_g$ accurately and with DT otherwise.
Theoretical results from \citet{ignatiadis2019covariate} show that, under certain regularity conditions and parameter estimation through sample splitting, EB has lower MSE compared to DT and \reg. 
Additionally, confidence intervals can be obtained for EB estimates, as detailed in \citet{armstrong2022robust}. We fdescribe their construction in \Cref{sec:additional-details-methods}.

\section{Results}
\label{sec:results}

In this section, we present the core set of results, conveying our key takeaways. 
We defer to \Cref{sec:add_additionalexp} for additional experiments, including a comparison with other baselines and other data modalities.

\subsection{Tasks and models}
\label{sec:taks_and_models}

We access the accuracy of LLMs across multiple-choice question-answering (MC QA) datasets, including
a subset of BIG-bench \citep{srivastava2022beyond},  
HellaSwag \citep{zellers2019hellaswag},
MedMCQA \citep{pal2022medmcqa}, 
MMLU \citep{hendrycks2020measuring}, 
PROST \citep{aroca2021prost},
Toxigen \citep{hartvigsen2022toxigen}, 
XNLI \citep{conneau2018xnli}, 
and XCOPA \citep{ponti2020xcopa}. 
Each dataset is analyzed independently, with subgroups defined by topics, domains, or tasks defined in the data.
Additionally, we also examine 
ANLI \citep{nie-etal-2020-adversarial}, 
ARC \citep{clark2018think}, 
MathQA \citep{amini-etal-2019-mathqa}, 
and Swag \cite{zellers2018swag}, which do not have pre-defined categories.
On these datasets, we form subgroups through unsupervised clustering of the text embeddings via \kmeans, choose the number via the silhouette score, and then aggregate smaller subgroups. 
Lastly, we also consider tasks involving text generation included in BIG-bench, SQuAD2.0 \citep{rajpurkar2018know}, and TriviaQA \citep{joshi2017triviaqa}.
The number of subgroups $G$ varies between 40 (PROST) and more than 200 (BIG-bench).

In terms of LLMs, we use 
Mistral-7B-Instruct-v0.2 \citep{jiang2023mistral} 
Google Gemma-2b \citep{team2024gemma}, 
Microsoft Phi-3-Mini-4K-Instruct \citep{abdin2024phi}, 
and OpenLLama-3B-v2 \citep{openlm2023openllama}.
All text embeddings are generated using an off-the-shelf BERT encoder \citep{devlin2018bert}. 

\subsection{Experimental setup}
\label{sec:expsetup}

To mimic what a practitioner would do, we consider each dataset separately but benchmark multiple LLMs at the same time. 
Thus, our evaluation process is as follows:
(i) \emph{Data sampling.}
Sample evaluation data from the entire dataset proportionally to subgroup sizes, ensuring that the smallest subgroup with at least 50 observations includes a minimum of $n_g=10$ observations in the sample. We also experiment with $\smash{n_g=10,20,50}$ for all subgroups, see \Cref{sec:app_more_llm_exp}. 
(ii) \emph{Model tuning and estimation.} 
Tune regression models via cross-validation, estimate the subgroup metrics $\hat\mu_g$ on the evaluation data using DT, \reg, and EB for all LLMs, and compute the corresponding confidence intervals. 
(iii) \emph{Repeated sampling and evaluation.} 
Repeat the process 1000 times and compute the MSE for each subgroup. 
``Ground truth'' subgroup metrics $\mu_g$ are estimated on the entire dataset.
We compare the methods by their average MSE as in \eqref{eq:evaluation_metric}. 
A lower average MSE indicates a better method. 

\paragraph{EB estimation.\hspace{-0.5em}}  Our implementation of the EB approach is based on the standard cross-fitting procedure of \citet{ignatiadis2019covariate} and is summarized in the following steps. 
(1) Estimate \(\hat{\sigma}^2_g\) for all \(g\)'s. 
(2) Split the sample into two folds, \(\mathcal{D}_1\) and \(\mathcal{D}_2\). 
(3) Fit \reg on \(\mathcal{D}_1\), obtain \reg estimates on  on \(\mathcal{D}_2\) and compute \(\hat{A}\). 
(4) Use these to obtain EB estimate for each \(g\) in \(\mathcal{D}_2\). 
(5) Repeat the process with the folds inverted to generate the EB estimates \(\{\hat{\mu}_g\}_{g=1}^G\).  

\subsection{Results}
\label{sec:results_llm}

\paragraph{Accuracy estimation in MC QA.\hspace{-0.5em}} 
\Cref{fig:results_main} gives a finegrained view of our results, showing the ratio of the MSE for the EB and \reg estimators compared to the DT estimator, across various models and datasets. 
The EB estimator consistently performs as well as or better than the two baselines, achieving on average 20\% and 30\% lower MSEs than \reg and DT respectively. 
\reg generally outperforms DT, although there are exceptions, e.g., BIG-bench, where most of the subgroups are large (and therefore DT estimates are precise).
On other datasets (e.g., ANLI), \reg outperforms DT, but EB typically still emerges as the most precise estimator.

\begin{figure}[t]
    \includegraphics[width=0.99\columnwidth]{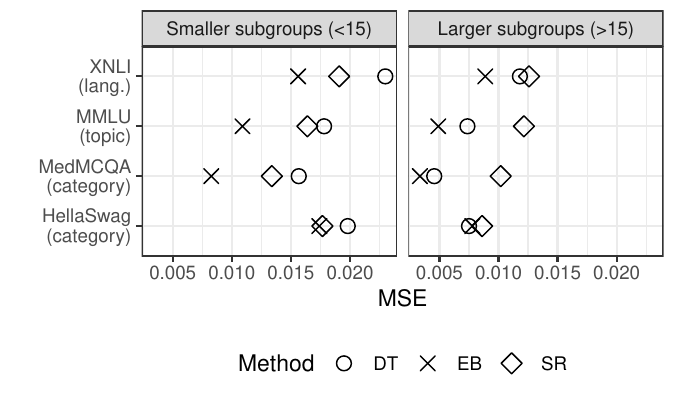}
    \caption{
        \textbf{Comparison of subgroup MSEs across methods.} 
        The plot compares the MSEs across all subgroups (LLM-domain pairs) across four datasets. \reg tends to perform better than DT on small subgroups but not always on larger ones. 
        EB performs better than both on either. An MSE$=0.01$ means that, on average, we have $|\hat\mu_g-\mu_g|=0.1$.  
    }
    \label{fig:llm_performance_dataset}
    \vspace{-1em}
\end{figure}

\Cref{fig:llm_performance_dataset} separates the MSE for larger ($>15$ questions) and smaller ($\leq 15$ questions) subgroups. 
We observe that DT generally provides more accurate estimates than \reg on the former due to small subgroup variances $\sigma^2_g$. 
In these cases, EB estimates align closely with those from DT.
For the smaller subgroups, \reg tends to present lower MSE than DT. 
In these cases, EB estimates mostly align with \reg's, resulting in improved precision compared to DT. 
Thus, by shrinking the estimates of \reg more heavily towards those of DT's when subgroup sizes are large (e.g., Big-bench) or when the regression fit is poor (e.g., MMLU, PROST), implying that $\hat{A}/\hat\sigma^2$ is large, EB yields more precise estimates compared than the baselines.

To determine the drivers of the precision gains of \reg over DT, we refit \reg removing groups of features one at a time, following the LOCO (Leave Out COvariates) approach of \citet{lei2018distribution}; we refer readers to \citet{verdinelli2023feature} for comparisons between this strategy and Shapely values \citep{lundberg2017unified}. 
Across all datasets, we find that removing model-specific intercepts as well as confidence scores leads to the largest increase in MSE, which we attribute to variations in the accuracy of the LLMs.
The removal of embeddings and task-related features do not appear to be strongly associated with the precision of the estimates.

Overall, we find that due to strong regularization, \reg estimates vary across LLMs but remain consistent across tasks for the same LLM. This strategy is effective only when subgroup performance is similar within a given LLM, aligning with our observation that most features are not highly informative. Considering also the sizes of the subgroups—typically large in BIG-bench and small in MMLU—this explains the performance differences between \reg and DT.
Lastly, benchmarking LLMs separately only slightly decreases the efficiency of EB, which still outperforms DT. 
Thus, even if only one LLM is evaluated, EB should be used.

\begin{figure}[t]
    \includegraphics[width=0.99\columnwidth]{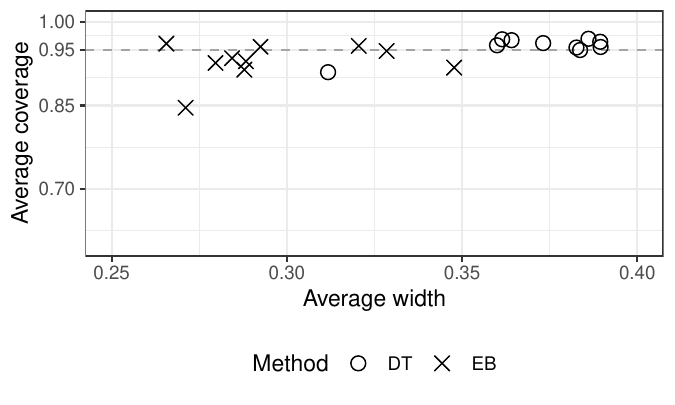}
    \caption{
        \textbf{Average coverage and width of 95\% confidence intervals} for DT and EB estimates of LLM accuracy across datasets. 
        EB intervals maintain high coverage and are generally narrower than those of DT. 
    }
    \label{fig:cis}
    \vspace{-1.25em}
\end{figure}

\paragraph{Coverage and width of confidence intervals.} 
We evaluate confidence intervals in terms of their coverage (i.e., the frequency with which they capture $\mu_g$) and their width, averaged across all subgroups. 
Since EB estimates are more precise than \reg, we focus on comparing DT and EB. 
Ideally, confidence intervals should achieve nominal coverage and be as narrow as possible. 
As shown in \Cref{fig:cis}, DT intervals consistently achieve nominal coverage (here $95\%$), while EB intervals also maintain high coverage, with most exceeding $\geq 90\%$. 
However, the average width of EB intervals is on average 20\% smaller than those of DT, indicating that EB estimates come with tighter confidence intervals.

\begin{table}[t]
    \vspace{-1em}
    \centering
    \begin{tabular}{cccc}
        \toprule
        Data/MSE$\times 10^3$ & \textbf{DT}  & \textbf{SR} & \textbf{EB} \\
        \midrule
        BIG-bench (\emph{exact match}) &  \textbf{1.0} & 19.5 &  1.6 \\
        SQuAD2.0 (\emph{F1 score}) & 246.0 & 178.0 & \textbf{109.5} \\
        Trivia QA (\emph{exact match}) & 12.2  & 12.8 &  \textbf{9.6}  \\
        \bottomrule
    \end{tabular}
    \caption{
        \textbf{Estimation method comparison on general tasks.} 
        The table compares the average MSEs of the estimates of all LLM performances across different datasets (with respective evaluation metrics) by the different methods. 
        EB consistently achieves similar or lower MSE than DT and \reg. 
    }
    \label{table:performance}
    \vspace{-1em}
\end{table}

\paragraph{Estimation for general tasks.\hspace{-0.5em}} 
Finally, we extend our comparison of estimation methods to tasks beyond MC QA.
\Cref{table:performance} summarizes the average MSE of each method on the BIG-bench, SQuAD2.0, and Trivia QA datasets.
While DT outperforms \reg on BIG-bench, \reg outperforms DT on SQuAD2.0. Their performance is similar on TriviaQA. 
EB remains the best-performing estimator on two datasets and its performance is close to that of DT on BIG-bench.

\section{Discussion}
\label{sec:discussion}
Our investigation contributes to a growing body of work on model evaluation in presence of resource constraints \citep{fogliato2024framework, zrnic2024active, herlihy2024structured}.
Our results show that an EB estimator, which combines DT and \reg estimates, consistently provides more precise estimates of subgroup-level model performances compared to DT and \reg alone. 
When subgroups are large, the estimates of EB align with those of DT. 
The uncertainty of EB estimates can also be effectively quantified via confidence intervals, which achieve near-nominal coverage and are consistently tighter than those derived for DT estimates (although see limitations in \Cref{sec:limitations}). 
Overall, EB proves to be a simple and useful method for model evaluation when estimating performance for lots of subgroups with (many or) few observations.

\bibliography{references}

\appendix

\clearpage

This serves as an appendix to the paper ``Precise Model Benchmarking with Only a Few Observations.''
The content of the appendix is organized as follows.

\section*{Organization} 

\begin{itemize}[leftmargin=5mm]
    \item 
    In \Cref{sec:additional-details-methods}, we provide additional details about the methods used in our paper. 
    In particular, we discuss in detail the following methods: 
    \begin{itemize}[leftmargin=5mm]
        \item \Cref{sec:synthetic-regression}: synthetic regression
        \item \Cref{sec:empirical-bayes}: empirical Bayes
    \end{itemize}
    \item In \Cref{sec:limitations}, we discuss the limitations of the methods. 
    \item In \Cref{sec:add_additionalexp}, we present additional experiments to further validate our results. 
    In particular, we show results on the following data types:
    \begin{itemize}[leftmargin=5mm]
        \item \Cref{sec:app_more_llm_exp}: additional LLMs
        \item \Cref{sec:vision_experiments}: computer vision tasks
        \item \Cref{sec:image-captioning}: image captioning task
        \item \Cref{sec:tabular-data}: tabular data
    \end{itemize}
\end{itemize}

\section{Additional Details on Methods}
\label{sec:additional-details-methods}

In this section, we provide additional details on the methods discussed in the main text and discuss other baselines. 
While the main body of the manuscript focuses on an essential comparison, here we also present methods proposed by previous work. 

\subsection{Synthetic regression (\reg)}
\label{sec:synthetic-regression}

The term ``synthetic'' is taken from the small area literature, where it is used to indicate estimators that leverage information from large subgroups about a target quantity to derive estimates on smaller subgroups under the assumption that these subgroups share similar characteristics associated with the target quantity \citep{ghosh1994small, rao2015small}. 

\paragraph{Implementation.\hspace{-0.5em}}
In principle, the synthetic regression approach can utilize any family of predictive models (e.g., boosted trees, linear regression, neural nets). 
We use a XGBoost regressor predictor to predict $Z_g$, tuning the number of trees and the depth via 2-fold cross-validation, stratifying by the task type. 
In the covariates $\tilde{X}_g$, we include (i) the values of the embeddings averaged across all observations in the subgroup, (ii) the average LLM confidence scores, (iii) intercepts for each model being evaluated and (iv) for each task. 
Other feature construction methods are also possible and might lead to better results; we focus on these regression features because they are intuitive and straightforward to obtain.

\paragraph{Structured regression.\hspace{-0.5em}}
A special case of \reg is \emph{structured regression}, which has recently been proposed by \citet{herlihy2024structured}. 
This approach involves fitting a Lasso regression model \citep{tibshirani1996regression} to minimize the following loss function:
\begin{equation}
    \label{eq:lasso}
     \argmin_{\beta_0 \in \RR, \beta_1 \in \RR^{p}} \sum_{g=1}^G \frac{1}{\sigma^2_g} (\beta^\top \tilde{X}_g - Z_g)^2 + \lambda \sum_{i=1}^p \vert \beta_i\vert,
\end{equation}
where $\beta = (\beta_0, \dots, \beta_p)^\top \in \mathbb{R}^{p}$. 
We include subgroup-level intercepts  in the feature set $\tilde{X}_g$ (e.g., on for each LLM-task pair), along with other explanatory features as in \reg (e.g., the embeddings). 
We tune $\lambda$ via cross-validation and estimate $\sigma^2_g$ using the plug-in estimator, e.g., $Z_g (1 - Z_g)/n_g$ for accuracy metrics. 
Note that if the number of features outnumbers the number of observations as it occurs in our experiments, we should expect that the predictions of this approach will converge to $Z_g$ as $\lambda \to 0^{+}$.
This is because the estimator in \eqref{eq:lasso} ``interpolates'' the training data and, among all such interpolators, it has the minimum $\ell_1$-norm; see, e.g., \citet{patil2022mitigating}.

\paragraph{Methods comparison.\hspace{-0.5em}} We distinguish \reg from structured regression to emphasize the fact that any modeling approach and feature construction can be used in the former. For example, we arbitrarily choose not to include subgroup-level intercepts in our regression model. 
Structured regression represents an alternative to our EB approach, where shrinkage is integrated directly into \reg instead of explicitly through the $\hat{A}$ and $\sigma^2_g$ parameters. 
Overall, we find that EB is more flexible because practitioners can use any predictive model and feature set, making it easier to implement. 
In practice, we will see that, when both approaches use Lasso regression, their performance is similar. 

\subsection{Empirical Bayes (EB)}
\label{sec:empirical-bayes}

The empirical Bayes estimator in our work is motivated by the following generative model \citep{armstrong2022robust}:
For $1\leq g \leq G$, assume that 
\begin{gather}
    \mu_g \mid X_g, \sigma^2_g \sim \mathcal{N}(f(X_g), A) 
    \label{eq:eb_model-2},\\
    \label{eq:eb_model-1}
    Z_g \mid \mu_g,X_g,\sigma_g^2 \sim \mathcal{N}(\mu_g,\sigma^2_g)
\end{gather}
where $A > 0$. 
Estimator \eqref{eq:eb} is derived from the posterior distribution of this data-generation mechanism. 
We recall it here:
\begin{equation}
    \label{eq:eb_estimation_appendix}
    \hat{\mu}_g=\hat{f}(X_g) + \frac{\hat{A}}{\hat{\sigma}_g^2+\hat{A}} \cdot (Z_g-\hat{f}(X_g)),
\end{equation}
where 
\begin{equation}
    \label{eq:hat_A}
    \hat{A}=\bigg( \frac{1}{G}\sum_{g=1}^G \epsilon_g^2-\hat\sigma_g^2 \bigg)_{+}.
\end{equation} 
Here, $\smash{\epsilon_g = Z_g-\hat{f}(X_g)}$, $\hat\sigma_g^2$ is estimated as in the \reg approach, and $(x)_{+}$ denotes the positive part of a real number $x$.

\paragraph{Boundary behavior.\hspace{-0.5em}} 
Observe the following two extreme cases:
\begin{itemize}[leftmargin=5mm]
    \item
    If $\hat\sigma^2_g\ll \hat{A}$, EB estimates will be close to DT.
    This case can occur when $n_g$ is large or \reg does not explain well the $\mu_g$'s.
    \item 
    If $\hat\sigma^2_g\gg \hat{A}$, EB estimates align with \reg.
    This case can occur when $Z_g$ has high variance. 
\end{itemize}

\paragraph{Confidence intervals.\hspace{-0.5em}} 
There are several methods for constructing confidence intervals for EB estimates \citep{laird1987empirical, rosenman2023combining}. 
We choose the approach of \citet{armstrong2022robust}, which is robust to violations of the assumption in \eqref{eq:eb_model-2}. 
More specifically, the confidence intervals are valid even when \eqref{eq:eb_model-2} is replaced by the following assumption: 
\begin{equation}
    \label{eq:eb_model-3}
    \mathbb{E}[(\mu_g-f(X_g))^2 \mid X_g,\sigma_g^2] = A.
\end{equation}
Their construction is fairly technical but has been implemented in existing software \citep{bowen2022multiple}.
Briefly, let 
\begin{align}
    \hat\kappa = \sum_{g=1}^G \frac{\epsilon_g^4 - 6\hat\sigma^2_g \epsilon_g^2 + 3 \hat\sigma_g^4}{\hat{A}^2}.
\end{align} 
Then the $1-\alpha$ confidence interval for $\mu_g$ is given by:
\begin{align}
    \hat\mu_g \pm \textrm{cva}_{\alpha}(\hat\sigma^2_g/\hat A, \hat\kappa) \cdot \frac{\hat{A}}{\hat\sigma^2_g + \hat A} \cdot \hat\sigma_g,
\end{align}
where the so-called critical value $\textrm{cv}_\alpha$ replaces the traditional percentile of the standard normal distribution in the typical intervals to account for the bias in the shrinkage \citep{armstrong2022robust}. 
Differently from the standard intervals that we construct for DT estimates, these intervals have an ``average coverage'' property:
While DT intervals cover each parameter with likelihood at least $1-\alpha$, the average coverage level of EB intervals across all $\mu_g$'s is at least $1-\alpha$. 

\paragraph{More background on composite estimators.\hspace{-0.5em}} 
Estimators of the form \eqref{eq:eb}, which combine DT with other estimates, are more broadly known as composite estimators \citep{rao2015small}.   
This class includes the classical James-Stein estimator \citep{james1992estimation}, which shrinks all $Z_g$ values towards the same chosen value, which is generally taken to be the global mean, e.g., see Section 2 in \citet{efron1975data}. 
Another estimator commonly used in small area estimation is the Fay-Herriot model, which assumes that $f(X)$ is linear \citep{fay1979estimates}. 
The version of EB that we study allows for arbitrary SR models and thus covers these cases. 
We expect the James-Stein estimator to perform similarly in cases where the regression employs strong regularization.

\section{Limitations of Methods}
\label{sec:limitations} 

Empirical Bayes methods are built on stronger assumptions than the standard direct estimator. 
These assumptions can allow us to increase the precision of the estimates and our experiments have shown that in general this occurs. 
However, this is not guaranteed and, even when it occurs, it may not always be desirable. 
For example, if one is interested in estimating only a subset of the $\mu_g$'s, then they should not seek to minimize \eqref{eq:evaluation_metric}. 
Another critical observation is the nature of the confidence intervals for EB estimates that we discuss. 
While standard intervals for DT are constructed to cover each $\mu_g$ with a given likelihood, EB intervals only have an average coverage guarantee. 
This should be noted when we aim to conduct hypothesis testing on the $\mu_g$'s using EB estimates. 
Practitioners that misunderstand this point might risk making erroneous inference and draw unsupported conclusions. 
In addition, the efficacy of EB over DT also hinges on how well the regression models $\mathbb{E}[Z_g \,|\, X_g]$. 
Therefore, feature construction and model selection are crucial. 
Our research has begun to explore these aspects, but further investigation is needed.

\section{Additional Experiments}
\label{sec:add_additionalexp}

In this section, we provide additional details on our main set of results and additional experiments to evaluate the performance of the methods.

\subsection{Additional experiments on LLMs}
\label{sec:app_more_llm_exp}

To generate the LLM confidence scores and associated predictions in all our experiments, we use the code in the lm-evaluation-harness repository by Eleuther AI \citep{eval-harness}. 
In this section, we review the results of additional experiments, including:
\begin{itemize}[leftmargin=5mm]
    \item An experiment where all subgroups have exactly $\smash{n_g\in\{10,20,50\}}$ observations (\Cref{sec:app_language_samesize}), including an analysis of how EB balances DT and SR estimates
    \item A comparison between EB and structured regression
\end{itemize}

\subsubsection{Results on all subgroups with equal size}\label{sec:app_language_samesize}
In this benchmarking exercise, we equalize the number of observations drawn from equal subgroups, setting either $\smash{n_g=10}$, $\smash{n_g=20}$, or $\smash{n_g=50}$. 
We drop all subgroups with less than $4 n_g$ observations. %
Thus, the experimental setup follows \Cref{sec:expsetup} with the exception of (i). 

The results from this experiment are summarized in \Cref{table:performance_allsubgroupssamesize}, which shows the mean relative efficiency of DT vs. SR or EB across datasets, which is defined as the ratio of the average MSE of \reg or EB estimates over the DT one. 
We observe that, as the size  of the subgroups increases, the gap in the precision of \reg and DT estimates shrinks. 
This occurs because DT estimates become more accurate. 
EB shows a similar trend that is nonetheless less remarked: 
Even for $n_g=50$, EB estimates have lower MSEs than those of DT. 
For this number of observations, 95\% confidence intervals for DT are fairly large (e.g., $\hat\mu_g\pm 0.08$ for $\mu_g=0.1$). Consequently, the gain in precision can result in considerable savings for practitioners. 

\begin{table}[!t]
    \centering
    \begin{tabular}{ccc}
        \toprule
        Size/Rel.\ eff.\   & \textbf{MSE SR / DT} & \textbf{MSE EB / DT} \\
        \midrule
        $n_g=10$ & 0.90 & 0.81   \\
        $n_g=20$ &  1.03  & 0.84 \\
        $n_g=50$ &  1.52  &  0.86 \\
        \bottomrule
    \end{tabular}
    \caption{
        \textbf{Comparison of estimation methods on LLM MC QA tasks with subgroups of equal size.} 
        The table compares the median over datasets of the relative efficiency (rel.\ eff.) of DT with respect to \reg and EB, namely the ratio of the average MSE of \reg or EB divided by the average MSE of DT when all subgroups have size $n_g$. 
        As $n_g$ increases, efficiency gains over DT decrease. 
    }
    \label{table:performance_allsubgroupssamesize}
\end{table}

\begin{figure*}[t]
    \centering
    \includegraphics[width=1.9\columnwidth]{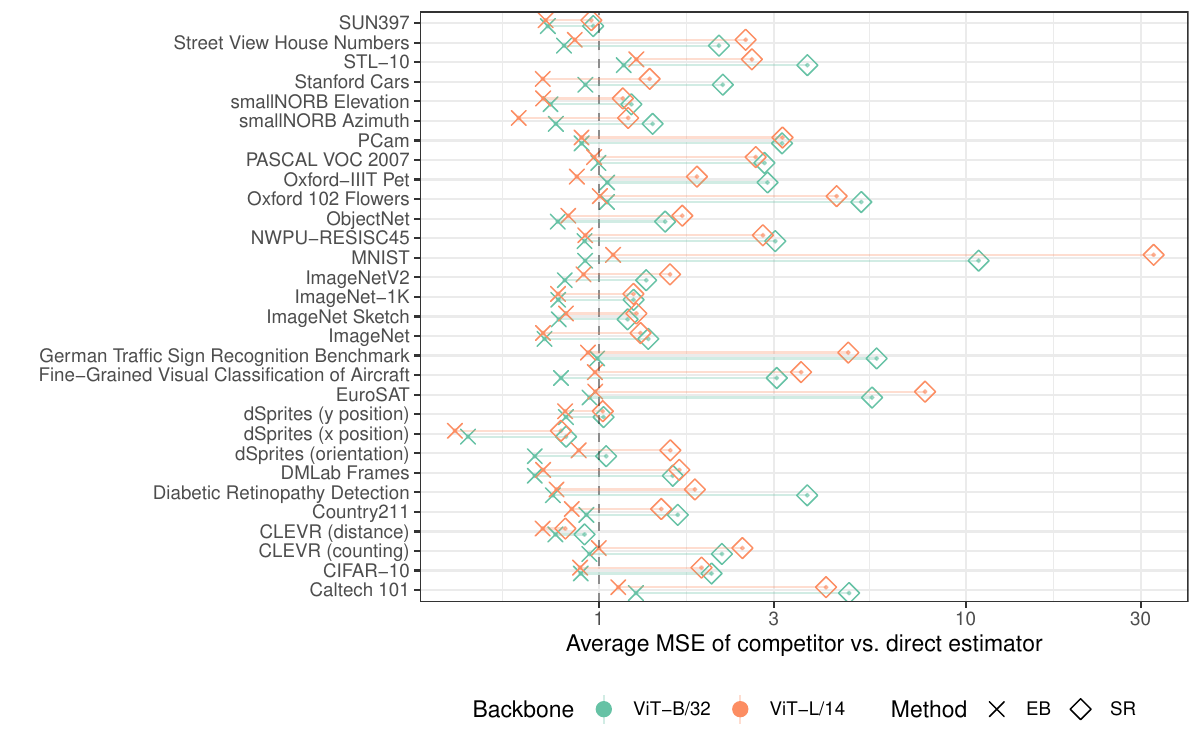}
    \caption{
        \textbf{Comparison of methods across datasets to estimate CLIP's zero-shot accuracies on subgroups of classification tasks.} 
        See the full list of tasks in \Cref{sec:vision_experiments}. 
        The observations correspond to the ratio between the average MSE of \reg or EB over the average MSE of DT estimates. 
        EB yields more precise estimates than \reg and DT across most datasets and models.
    }
    \label{fig:vision_performance_dataset}
\end{figure*}

\subsubsection{Comparison with structured regression}

We assess the precision of EB estimates against those of the structured regression estimator described in \Cref{sec:additional-details-methods}. 
\Cref{table:performance_additionalcomparison} shows the corresponding results, for both the setting where subgroup sizes are proportional to the original data ($n_g=10$ for the smallest group) and where all $n_g=20$ and all subgroups have the same size. 
We find that EB beats structured regression across almost all datasets. 
The relative efficiency of their estimates, however, reveals that the methods have comparable efficiency. 
Thus, practitioners should choose the method that is easiest to implement.

\subsection{Experiments on computer vision tasks}
\label{sec:vision_experiments}

\begin{table}[!t]
    \centering
    \resizebox{\columnwidth}{!}{%
    \begin{tabular}{ccc}
        \toprule
        Setting/Rel.\ eff.\ & \textbf{EB} & \textbf{Str.\ Reg.\ } \\
        \midrule
        $\min_{g\in [G]} n_g=10$, prop. & 0.62 (0.14) & 0.68 (0.34) \\
        $n_g=20$, equal & 0.64 (0.15) & 0.73 (0.47) \\
        \bottomrule
    \end{tabular}
    }
    \caption{
        \textbf{Estimation method comparison on LLM MC QA tasks of EB with other baselines.} 
        The table compares the mean (standard deviation in parentheses) over datasets of the relative efficiency (rel. eff.) of DT with respect to three methods: EB and structured regression. 
    }
    \label{table:performance_additionalcomparison}
\end{table}

We explore the performance of the methods on a series of computer vision classification tasks. 
We consider most of the datasets and tasks included in the LAION CLIP benchmark \citep{laion_clip_benchmark}, including 
Caltech 101 \citep{fei2004learning}, 
Stanford Cars \citep{KrauseStarkDengFei-Fei_3DRR2013}, 
CIFAR-10 \citep{krizhevsky2009learning}, 
CIFAR-100 \citep{krizhevsky2009learning}, 
CLEVR (distance and count) \citep{johnson2017clevr}, 
Describable Text Features \citep{cimpoi14describing}, 
DR Detection \citep{diabetic-retinopathy-detection}, 
DMLab Frames \citep{zhai2019visual}, 
EuroSAT \citep{helber2019eurosat}, 
FGVC aircraft \citep{maji13fine-grained}, 
Oxford 102 Flower \citep{nilsback2008automated}, 
GTSRB \citep{stallkamp2011german}, 
ImageNet-A \citep{hendrycks2021nae}, 
ImageNet-R \citep{hendrycks2021many}, 
ImageNet-1K \citep{imagenet15russakovsky}, 
ImageNet Sketch \citep{wang2019learning}, 
ImageNetV2 \citep{recht2019imagenet}, 
KITTI Distance \citep{Geiger2013IJRR}, 
MNIST \citep{deng2012mnist}, 
ObjectNet \citep{barbu2019objectnet}, 
Oxford-IIIT Pet \citep{parkhi2012cats}, 
PASCAL VOC 2007 \citep{pascal-voc-2007}, 
PCam \citep{veeling2018rotation}, 
Rendered SST-2 \citep{socher2013recursive}, 
NWPU-RESISC45 \citep{cheng2017remote}, 
SmallNorb (Azimuth and Elevation) \citep{lecun2004learning}, 
STL-10 \citep{coates2011analysis}, 
SUN397 \citep{Xiao:2010}, 
Street View House Numbers \citep{netzer2011reading}. 

We experiment with predictions generated by CLIP models that employ ViT-B/32 and ViT-L/14 as vision backbones \citep{ilharco_gabriel_2021_5143773, schuhmann2022laionb, radford2021learning} in the zero-shot setting. 
We form task subgroups via \kmeans as in \Cref{sec:taks_and_models} and sample $n_g=20$ observations from all subgroups; we found that sampling proportionally led to only a couple of small groups and many large ones. 
Our goal is to estimate the accuracy of the models on the subgroups.
The experimental evaluation setup is analogous to what we described in \Cref{sec:expsetup}. 
The setup of \reg also remains unchanged, with the exception that we run the model on the ViT-B-32's image embeddings instead of the text embeddings. 

\Cref{fig:vision_performance_dataset} shows the results of this set of experiments. 
We find that \reg beats DT on some datasets but severely underperforms on others. 
This is explained by the fact that (most often) the \reg estimates are heavily regularized and predictions are close across subgroups. As we saw in the case of LLMs, this strategy improves over DT on datasets where subgroup performances are similar (e.g., dSprites (x position)) yet it miserably fails when subgroup performance are far apart (e.g., in MNIST). 
In these cases, we observe that the EB estimates align closely those from DT, as a result of a large ratio $\hat A / \hat\sigma^2$. 
Despite the overall underperformance of \reg, we find that EB still performs similarly to DT and always better than \reg. 

\subsection{Experiment on image captioning}
\label{sec:image-captioning}

We assess the performance of EB on evaluating a BLIP \citep{DBLP:journals/corr/abs-2201-12086} image captioning model on a test split \citep{Karpathy_2015_CVPR} of 5000 images from the COCO Captions dataset \citep{10.1007/978-3-319-10602-1_48, DBLP:journals/corr/ChenFLVGDZ15}, where each of the test images includes 5 reference captions.  
We use the BLIP model to generate a new caption for each image in the test set and then evaluate the results using the CLAIR score \cite{chan-etal-2023-clair}.  
This score uses an LLM (in our experiments we use a Mistral 7B model \cite{jiang2023mistral}) to predict a similarity score between the generated caption and a set of reference captions.
For each test image, we average the CLAIR score over the set of reference captions to obtain a single score per image.

To evaluate our approach in this setting, we form subgroups by first generating image embeddings for each image using CLIP \cite{radford2021learning} and clustering the embeddings via \kmeans where the silhouette score was used to choose the number of clusters (about $40$ in our test case) as in the previous experiments.
The rest of the experimental setup also follows \Cref{sec:expsetup}.

\begin{figure}[!t]
    \centering
    \includegraphics[width=0.99\columnwidth]{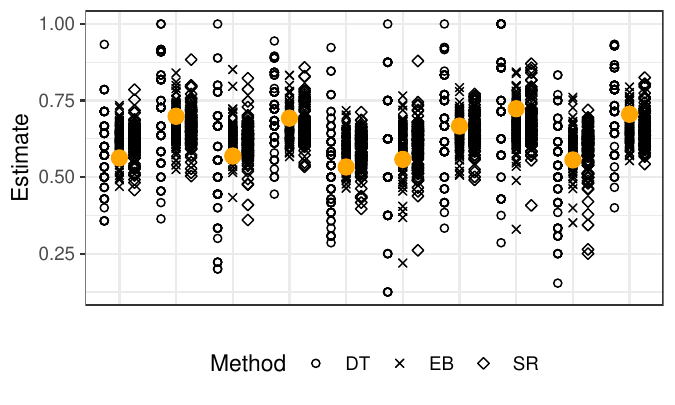}
    \caption{
        \textbf{Subgroup estimates of CLAIR scores across 10 subgroups in VQA data.} 
        The dots in \textcolor{orange}{orange} correspond to ground truth values, while those in black indicate the estimates obtained over $100$ random draws from the subgroups with the different approaches. 
    }
    \label{fig:captioning_results}
\end{figure}

We find that both \reg and EB substantially outperform DT, achieving average MSEs that are 70\% and 80\% lower than those of DT, respectively. 
\Cref{fig:captioning_results} shows the estimates across 10 randomly sampled subgroups in the data. 
We observe that, due to the small subgroup sizes in the evaluation data, the accuracy estimates of DT have high variance. 
Both those of DT and of EB have significantly less variance and, as our empirical results show, also happen to have little bias for most of the subgroups. 

\subsection{Experiments on tabular data}
\label{sec:tabular-data}

We present results for the subgroup performance estimation methods on a host of binary classification tasks commonly used in the algorithmic fairness literature. 
This includes five prediction tasks on $2018$ ACS folkstable data from New York \citep{ding2021folktables}, the COMPAS dataset \citep{Angwin2016}, and student performance datasets in math and Portuguese \citep{cortez2008using}. 
Similar to \citet{herlihy2024structured}, we define subgroups by intersections of protected attributes like race or ethnicity, age, and sex.
We then retain only those with at least $20$ observations for subgroup performance estimation.

\begin{figure}[!t]
  \centering
  \includegraphics[width=0.99\columnwidth]{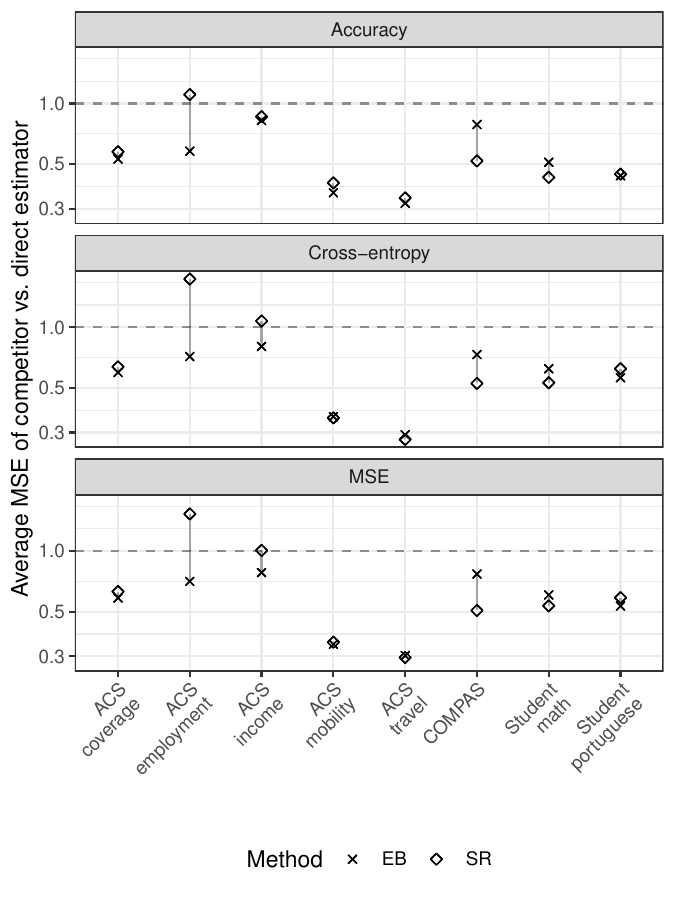}
   \caption{
       \textbf{Comparison of methods across tabular datasets.} 
       The plot shows the ratio of the MSE of the estimates of the LLM metric on the subgroups obtained through SR and EB (the competitors) over the MSE of those generated by DT. 
       Lower values indicate more precise estimates compared to DT. 
       As metrics, we consider accuracy, cross-entropy, and MSE or Brier score.
   }
   \label{fig:results_tabular}
\end{figure}

Predictions on tabular datasets are generated using XGBoost \citep{chen2016xgboost}. 
Each model is trained with 3-fold cross-validation on 50\% of the respective dataset. 
The other 50\% of the examples are retained for our evaluation set using the same setup of \Cref{sec:methods}. 

\Cref{fig:results_tabular} shows the ratio of the MSE of EB and \reg estimates to the DT estimates across all tabular datasets.
Similar to our observations in the LLM context, the EB estimator provides estimates that are just as precise or more precise than SR and DT estimators for six out of the eight prediction tasks. 
The EB estimator's advantage over SR is especially pronounced for the ACS Employment and Income datasets for which SR struggles to outperform the DT baseline. 
EB and \reg performances are virtually identical for the ACS mobility and travel tasks, implying that \reg explains little to no additional variance in these cases.
In the context of the COMPAS and Student Math datasets, the SR estimator provides better estimates than the EB estimator. 
Both of these datasets have less than 10 distinct subgroups which may lead to poor fits in computing $\hat{f}(X)$ and $\hat{A}$. 
Nevertheless, both \reg and EB estimators consistently outperform the DT baseline.

We thus conclude that, similar to the results in the main body of the paper, the EB estimator for tabular data provides more precise subgroup performance estimates than DT and also \reg when there are enough distinct subgroups. 

\end{document}